%% file: main.tex
\documentclass[10pt,twocolumn,letterpaper]{article}

\usepackage{cvpr}              
\usepackage{multirow}
\usepackage[accsupp]{axessibility} 
\usepackage[table]{xcolor}   
\definecolor{rowgray}{RGB}{242,242,242} 
\input{preamble}
\definecolor{cvprblue}{rgb}{0.21,0.49,0.74}
\usepackage[pagebackref,breaklinks,colorlinks,allcolors=cvprblue]{hyperref}
\usepackage{marvosym}
\def\paperID{***} 
\def\confName{CVPR}
\def\confYear{2026}

\title{Text–Image Conditioned 3D Generation}

\author{
Jiazhong Cen\textsuperscript{\rm 1}\footnotemark[2] \quad
Jiemin Fang\textsuperscript{\rm 2 \Letter} \quad
Sikuang Li\textsuperscript{\rm 1}\footnotemark[2] \quad
Guanjun Wu\textsuperscript{\rm 3}\footnotemark[2] \quad
Chen Yang\textsuperscript{\rm 2} \quad
Taoran Yi\textsuperscript{\rm 3}\footnotemark[2] \\
Zanwei Zhou\textsuperscript{\rm 1}\footnotemark[2] \quad
Zhikuan Bao\textsuperscript{\rm 2} \quad
Lingxi Xie\textsuperscript{\rm 2} \quad
Wei Shen\textsuperscript{\rm 1 \Letter} \quad
Qi Tian\textsuperscript{\rm 2} \\
$^{1}$MoE Key Lab of Artificial Intelligence, AI Institute, \\
School of Computer Science, Shanghai Jiao Tong University \\
$^{2}$Huawei Inc. \\
$^{3}$Huazhong University of Science and Technology \\
{\tt\small jaminfong@gmail.com, wei.shen@sjtu.edu.cn}
}

\begin{document}
\maketitle

{
\renewcommand{\thefootnote}{\fnsymbol{footnote}}
\footnotetext[2]{Work done during internship at Huawei.}
\footnotetext[0]{\hspace{-4pt}$^\text{\Letter}$Corresponding authors.}
}

\input{0_abstract}
\input{1_intro}
\input{2_related}
\input{3_method}
\input{4_experiments}

\input{5_conclusion}

\section*{Acknowledgements}
This work was supported by the NSFC under Grant 62322604 and 62576207. We thank the anonymous reviewers for their valuable feedback and suggestions.
{
    \small
    \bibliographystyle{ieeenat_fullname}
    \bibliography{main}
}

\input{X_suppl}

\end{document}

%% file: 0_abstract.tex
\begin{abstract}
High-quality 3D assets are essential for VR/AR, industrial design, and entertainment, motivating growing interest in generative models that create 3D content from user prompts. Most existing 3D generators, however, rely on a single conditioning modality: image-conditioned models achieve high visual fidelity by exploiting pixel-aligned cues bhaodeut suffer from viewpoint bias when the input view is limited or ambiguous, while text-conditioned models provide broad semantic guidance yet lack low-level visual detail. This limits how users can express intent and raises a natural question: can these two modalities be combined for more flexible and faithful 3D generation? Our diagnostic study shows that even simple late fusion of text- and image-conditioned predictions outperforms single-modality models, revealing strong cross-modal complementarity. We therefore formalize \textbf{Text–Image Conditioned 3D Generation}, which requires joint reasoning over a visual exemplar and a textual specification. To address this task, we introduce \textbf{TIGON}, a minimalist dual-branch baseline with separate image- and text-conditioned backbones and lightweight cross-modal fusion. Extensive experiments show that text–image conditioning consistently improves over single-modality methods, highlighting complementary vision–language guidance as a promising direction for future 3D generation research. Project page: \url{https://jumpat.github.io/tigon-page}
\end{abstract}




%% file: 1_intro.tex
\section{Introduction}
\label{sec:intro}

Generating high-quality 3D assets has attracted increasing attention due to its importance for downstream applications such as virtual reality, industrial design, and embodied AI. A particularly promising line of work~\cite{trellis, hunyuan3d, direct3ds2, triposr, step1x, clay} learns to generate 3D objects from either an input image or a text description. However, relying on a single conditioning modality can limit the flexibility of 3D generation.

\begin{figure}
    \centering
    \includegraphics[width=\linewidth]{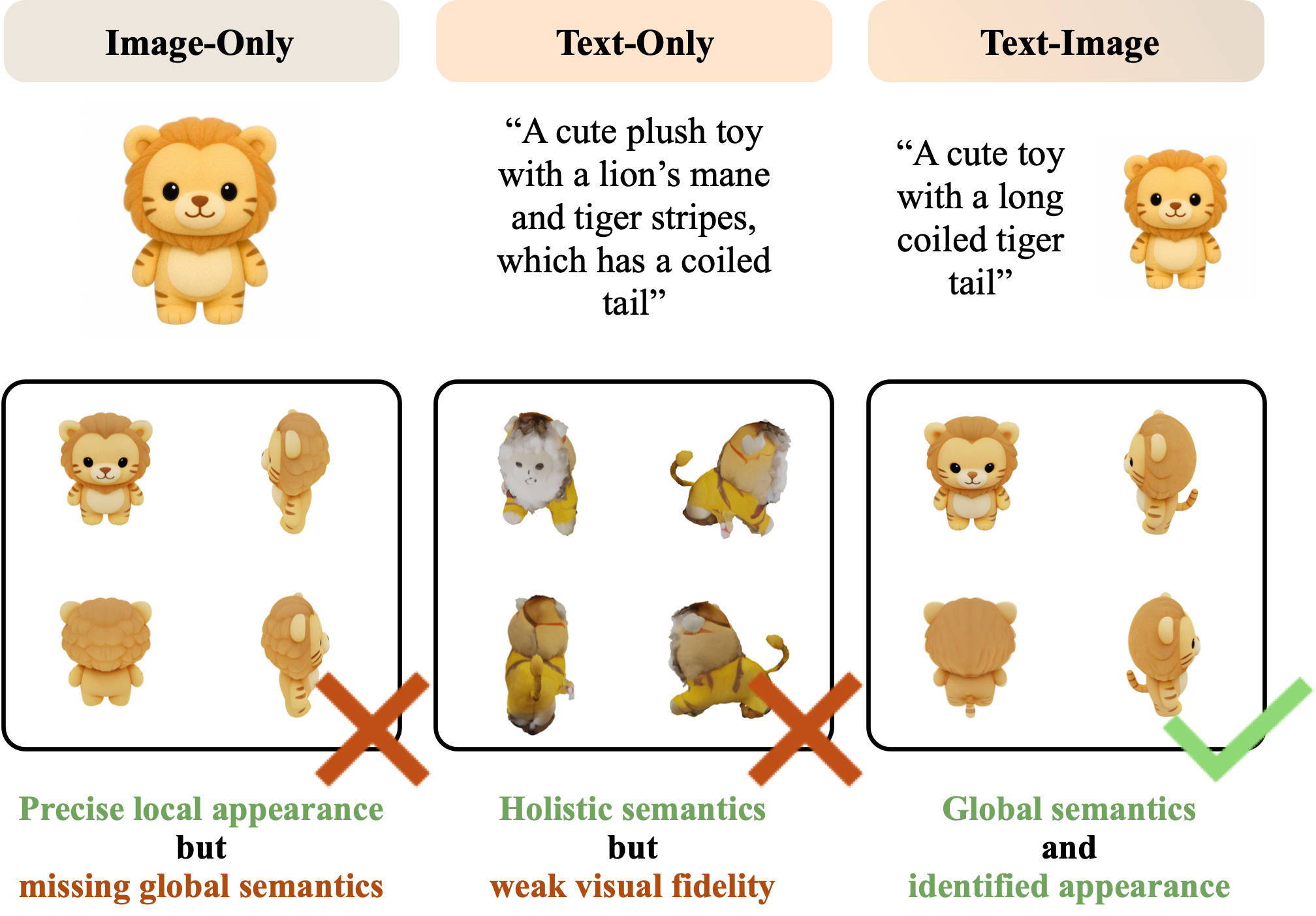}
    \caption{Single-modality conditioning has limitations in satisfying user intent.
    Image-only conditioning captures local appearance but omits unobserved regions; text-only conveys semantics but lacks visual fidelity.
    In contrast, joint text–image conditioning produces 3D assets that are both semantically aligned with the description and faithful to the reference appearance.}
    \label{fig:teaser}
\end{figure}

As shown in~\cref{fig:teaser}, when a user specifies a 3D object with an example image, image-conditioned 3D generation can preserve local appearance but is highly sensitive to viewpoint informativeness: occlusions, atypical views, or incomplete object coverage force the model to hallucinate under-constrained regions, causing the generated 3D asset to deviate from the intended semantics. In contrast, text-conditioned 3D generation is semantically reasonable but lacks concrete visual constraints, so the output may roughly match the prompt while exhibiting poor visual quality.

These observations raise a natural question of whether image and text conditions can provide complementary information for more flexible 3D generation. Intuitively, images anchor the result to the actual observed view, supplying reliable geometry and appearance cues, while text can specify additional semantics to disambiguate unobserved regions (\emph{e.g.}, ``with a long coiled tiger tail''). In a diagnostic study, we find that conditioning on a low-information view degrades performance, but adding a textual description and fusing the image- and text-conditioned predictions noticeably recovers quality. This motivates us to move beyond single-modality settings and introduce \textbf{Text--Image Conditioned 3D Generation}\footnote{Here we focus on the \emph{native} 3D generation setting, where the model directly generates a 3D representation under joint text--image conditioning.}, which requires the 3D generator to jointly reason over the visual exemplar and the textual specification, and to generate a consistent 3D asset that is simultaneously faithful to the image-conditioned appearance/geometry and aligned with the text-defined semantics.

To address this task, we propose a strong yet minimalist baseline named \textbf{TIGON}. It adopts a dual-branch design that retains two modality-specialized DiT backbones and couples them via two lightweight fusion mechanisms: (i) cross-modal linear bridges for bidirectional feature sharing (early fusion), and (ii) step-wise prediction averaging along the denoising trajectory (late fusion). This design prevents either branch from shouldering the cross-modal domain gap, preserves their original single-modality ability, and enables free-form conditioning. Extensive experiments show that TIGON delivers more flexible 3D generation.

Our contributions are summarized as follows: (1) We identify and empirically diagnose the limitations of existing single-modality 3D generation methods. (2) We show that image and text provide complementary conditioning signals and, motivated by this, introduce the task of text–image conditioned 3D generation. (3) We propose TIGON, a simple yet effective baseline method that leverages modality-specific backbones with lightweight cross-modal fusion. (4) We conduct extensive experiments to demonstrate that TIGON achieves more robust and flexible 3D generation.

%% file: 2_related.tex
\section{Related Work}
\label{sec:related}

\paragraph{3D Generation with 2D Generative Models.}
SDS-based text-to-3D methods optimize differentiable 3D representations (\emph{e.g.}, NeRFs~\cite{nerf}) using frozen 2D diffusion priors, as pioneered by DreamFusion~\cite{dreamfusion} and SJC~\cite{sjc}. Later works improve resolution, geometry--appearance disentanglement, optimization stability, and multi-view consistency, while recent hybrids with 3D Gaussian Splatting or native 3D diffusion further boost efficiency and fidelity~\cite{consistent3d,magic3d,fantasia3d,prolificdreamer,li2024connecting,kwak2024geometry,dreamcraft3d,gaussiandreamer,dreamgaussian,liang2024luciddreamer,li2023instant3d}. A parallel line studies image-conditioned 3D generation and reconstruction with 2D generative priors, either by directly optimizing 3D representations or by first generating consistent multi-view images and then reconstructing 3D content~\cite{realfusion,neurallift360,makeit3d,liu2023zero,weng2023consistent123,liu2024one,liu2023one,wonder3d,syncdreamer,xu2024instantmesh,li2023instant3d}. Despite strong progress, reliance on 2D priors often limits 3D consistency, motivating native 3D generative models and cross-modal latent modeling~\cite{michelangelo}.

Among prior works, TICD~\cite{ticd} and FlexGen~\cite{flexgen} are most related to ours. TICD augments SDS-based text-to-3D with an image diffusion prior, while FlexGen jointly conditions on text and image but focuses on 2D multi-view generation rather than native 3D synthesis. In contrast, we study \textbf{native} 3D generation under joint text-image conditioning.

\paragraph{Native 3D Generative Models.}
Unlike SDS-based pipelines that rely on 2D generators at test time, native 3D generative models operate directly on 3D representations such as point clouds, meshes, voxels, 3D Gaussians~\cite{3dgs,tang2024lgm}, and neural fields~\cite{rf, nerf, mipnerf, lan2024ln3diff}. Early works~\cite{nichol2022point, vahdat2022lion} introduce fast point-cloud synthesis via latent diffusion. Later models such as 3DShape2VecSet~\cite{zhang20233dshape2vecset} improve geometry-aware latents for diffusion training. Moreover, scalable voxel-based methods~\cite{ren2024xcube}, octree-based models~\cite{xiong2025octfusion} and 3DGS-based models~\cite{yang2024atlas} improve generation resolution and efficiency. 

More recent studies scale data and model capacity for high-quality asset synthesis. Some focus on geometry-only generation for detailed meshes~\cite{chen2025ultra3d,direct3d,direct3ds2,li2025sparc3d,ye2025hi3dgen}, while others target fully textured 3D assets~\cite{clay,3dtopiaxl,hunyuan3d2025hunyuan3d2.1,lai2025hunyuan3d2.5,step1x,zhou2025few}. Among the latter, TRELLIS~\cite{trellis} introduces a sparse, structured latent representation that enables high-fidelity 3D generation via a two-stage pipeline, and UniLat3D~\cite{unilat} provides a unified-latent variant with a simplified sampling process. Together, these advances establish native 3D diffusion models as strong alternatives to 2D diffusion-based pipelines. 
However, most existing approaches assume single-modality conditioning (either image or text), which limits the flexibility and expressiveness of user instructions. In this work, we study this limitation and introduce the task of text–image conditioned 3D generation.


\paragraph{Multimodal-Conditioned Generation.}
Multimodal conditioning has proven effective in text, image, and video generation. Multimodal LLMs that ingest both language and visual (or even 3D) inputs~\cite{llava,qwenvl,instructblip,internvl,3dllm,pointllm,ll3da} enable grounded reasoning and control, while text-to-image diffusion benefits from additional visual controls (edges, depth, pose, layout) and joint text-image inputs~\cite{controlnet,t2i,ipadapter,qwenimage,bagel,instructp2p} for better alignment with user intent. In video generation, combining a reference frame with textual guidance~\cite{makeitmove,wan} yields more consistent and controllable outputs. These successes motivate our exploration of whether joint text-image conditioning can similarly endow 3D generators with complementary strengths beyond single-modality conditioning.

%% file: 3_method.tex
\section{Preliminaries}
\label{sec:prelim}

\label{sec:trellis}
TRELLIS~\cite{trellis} is an effective 3D generator that learns rectified-flow models~\cite{lipman2022flow,esser2024scaling} for a geometry latent (which voxels are active) and an appearance latent defined on those active locations.

\paragraph{Geometry Generation.}
A 3D object is voxelized into a sparse set of occupied points $\{\mathbf{p}_i\}_{i=1}^N$, $\mathbf{p}_i \in \mathbb{R}^3$. A geometry VAE compresses and reconstructs this set:
\begin{equation}
    \mathbf{z}_{\text{geo}} = \mathcal{E}_{\text{geo}}(\{\mathbf{p}_i\}), 
    \qquad
    \{\mathbf{p}_i\} = \mathcal{D}_{\text{geo}}(\mathbf{z}_{\text{geo}}).
    \label{eq:trellis-geo}
\end{equation}
A rectified-flow model $\mathcal{F}_{\text{geo}}$ parameterizes a time-conditioned velocity field that transports Gaussian noise toward $\mathbf{z}_{\text{geo}}$ under image or text conditioning.

\paragraph{Appearance Generation.}
On the occupied positions $\{\mathbf{p}_i\}$, TRELLIS aggregates multi-view image features
\[
\mathbf{F} = \{(\mathbf{p}_i, \mathbf{f}_i)\}_{i=1}^N, \quad \mathbf{f}_i \in \mathbb{R}^d,
\]
extracted by a 2D vision encoder such as DINOv2~\cite{dinov2}, and encodes them into a Structured LATent (SLAT) with an appearance VAE:
\begin{equation}
    \mathbf{z}_{\text{SLAT}} = \mathcal{E}_{\text{app}}(\mathbf{F})
    = \{(\mathbf{p}_i, \mathbf{z}_i)\}_{i=1}^N.
    \label{eq:trellis-slat}
\end{equation}
A second rectified-flow model $\mathcal{F}_{\text{app}}$ is learned for $\mathbf{z}_{\text{SLAT}}$, and the appearance decoder produces the final 3D representation
\begin{equation}
    \mathcal{O} = \mathcal{D}_{\text{app}}(\mathbf{z}_{\text{SLAT}}),
\end{equation}
where $\mathcal{O}$ can be mesh, 3DGS~\cite{3dgs}, or radiance fields~\cite{rf}.

\paragraph{Sampling with Rectified Flow.}
For each latent $\ell \in \{\text{geo}, \text{SLAT}\}$ with decoder $\mathcal{D}_\ell \in \{\mathcal{D}_{\text{geo}}, \mathcal{D}_{\text{app}}\}$ and a chosen condition $\mathbf{c}$ (image $\mathbf{I}$ or text $\mathbf{T}$), inference integrates the rectified-flow ODE from noise to data. Let $1 = t_0 > t_1 > \cdots > t_K = 0$ be a fixed schedule and initialize
$\tilde{\mathbf{z}}_{\ell, t_0} \sim \mathcal{N}(\mathbf{0}, \mathbb{I}^{c})$, where $c$ is the latent dimension. At step $k$,
\begin{equation}
\begin{aligned}
        &\mathbf{v}_{\ell, k}
    = \mathcal{F}_{\ell}\!\left(\tilde{\mathbf{z}}_{\ell, t_k},\, t_k;\, \mathbf{c}\right),\\
    &\tilde{\mathbf{z}}_{\ell, t_{k+1}}
    = \tilde{\mathbf{z}}_{\ell, t_k}
      - (t_k - t_{k+1})\, \mathbf{v}_{\ell, k},
\end{aligned}
\end{equation}
yielding $\tilde{\mathbf{z}}_{\ell, 0}$ after $K$ steps.
Decoding $\tilde{\mathbf{z}}_{\text{geo}, 0}$ with $\mathcal{D}_{\text{geo}}$ produces the activated voxels, and decoding
$\tilde{\mathbf{z}}_{\text{SLAT}, 0}$ with $\mathcal{D}_{\text{app}}$ gives the final 3D output $\mathcal{O}$.

\paragraph{UniLat3D.} Based on TRELLIS, UniLat3D~\cite{unilat} offers a more convenient single-stage 3D generator. Given view-aggregated features $\mathbf{F}$, it encodes them as
\begin{equation}
    \mathbf{z}_{\text{uni}} = \mathcal{E}_{\text{uni}}(\mathbf{F}), \quad
    \mathbf{z}_{\text{uni}} \in \mathbb{R}^{16 \times 16 \times 16 \times c},
\end{equation}
and directly decodes to a 3D output $\mathcal{O} = \mathcal{D}_{\text{uni}}(\mathbf{z}_{\text{uni}})$.
A single rectified-flow model predicts $\tilde{\mathbf{z}}_{\text{uni}}$ during generation. We adopt UniLat3D for its simplicity and compatibility with our conditioning study.

\section{Text-Image Conditioned 3D Generation}
\label{sec:comp}

As discussed in~\cref{sec:intro}, image-conditioned 3D generation is vulnerable to viewpoint bias, while text-conditioned generation benefits from comprehensive semantics but lacks the visual cues needed for high-fidelity synthesis. In this section, we first empirically diagnose this limitation, then show that even a simple late fusion of image- and text-conditioned predictions yields noticeable gains, revealing clear complementarity between the two modalities and motivating the task of text–image conditioned 3D generation.

\begin{figure}
    \centering
    \includegraphics[width=0.9\linewidth]{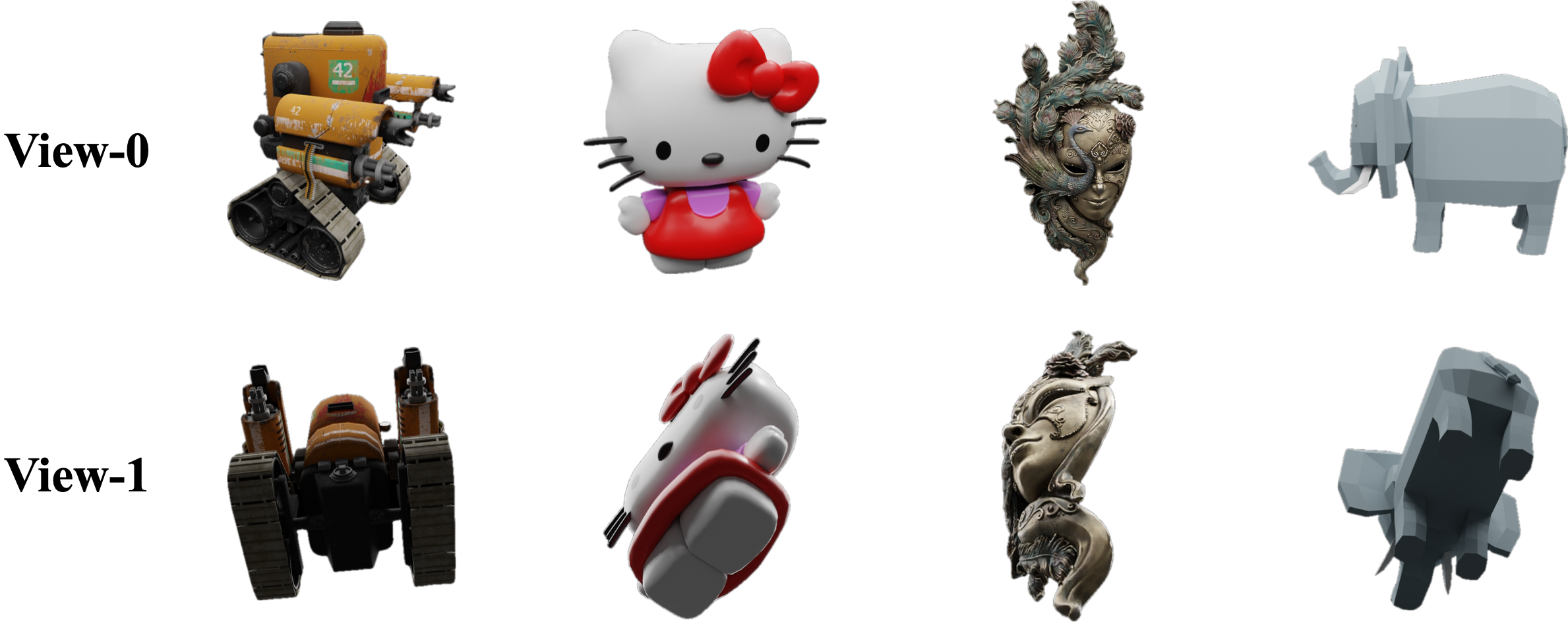}
    \caption{Reference views used in our diagnostic study. Moving from \emph{View-0} to \emph{View-1} reduces observable cues and creates a lower-information setting. Under this shift, single-modality baselines exhibit a marked performance drop.}
    \label{fig:viewpoint}
\end{figure}

\paragraph{Limitations of Single-Modality Conditioning.}
\cref{tab:view_0_res} reports the performance of representative 3D generation models on Toys4K under two viewpoint configurations\footnote{Please refer to~\cref{sec:eval_metric} for details about the metrics used in~\cref{tab:view_0_res}.}. As illustrated in Fig.~\ref{fig:viewpoint}, \textit{View-0} is a frontal view with rich semantics and clear local details, whereas \textit{View-1} is a low-angle view providing much weaker cues. This change alone leads to substantial degradation: TRELLIS degrades from $56.08$ FD$_{\text{DINOv2}}$ under \textit{View-0} to $143.58$ under \textit{View-1}, indicating strong dependence on viewpoint completeness. Moreover, the text-only counterparts perform even worse in visual alignment (\emph{e.g.}, UniLat3D reaches only $154.88$ FD$_{\text{DINOv2}}$), confirming that text priors alone are insufficient to recover fine-grained visual details.

\begin{table}[t]
\centering
\small
\setlength{\tabcolsep}{4.5pt}
\caption{Performance of existing methods on the Toys4K dataset under different conditioning signals. `GS' denotes that the 3D representation is 3DGS.}
\label{tab:view_0_res}
\begin{tabular}{lccc}
\toprule
\textbf{Model} & \textbf{Cond.} & \textbf{CLIP$\uparrow$} & \textbf{FD$_{\text{DINOv2}}\downarrow$} \\
\midrule
TripoSR                 & View-0 & 88.67 & 269.58 \\
Step1X-3D$^{\dagger}$   & View-0 & 89.99 & 152.69 \\
Hunyuan3D-2.1$^{\dagger}$ & View-0 & 89.87 & 114.64 \\
TRELLIS (GS)               & View-0 & 92.88 & 56.08  \\
UniLat3D (GS)                & View-0 & 93.34 & 47.41 \\
\midrule
TripoSR                 & View-1 & 79.40 & 804.18   \\
Step1X-3D$^{\dagger}$   & View-1 & 80.47 & 562.84    \\
Hunyuan3D-2.1$^{\dagger}$ & View-1 & 85.33 & 229.36   \\
TRELLIS (GS)                  & View-1 & 88.16 & 143.58 \\
UniLat3D (GS)                 & View-1 & 89.03 & 125.93    \\
\midrule
TRELLIS (GS)                  & Text   & 86.30    & 148.21    \\
UniLat3D (GS)                 & Text   & 86.14    & 154.88 \\
\midrule
SimFusion (GS; Ours)        & View-1 + Text & \textbf{90.64} & \textbf{82.40} \\
\bottomrule
\end{tabular}
\vspace{2pt}
\begin{flushleft}
\footnotesize
$^{\dagger}$ Using non-public training data.
\end{flushleft}
\end{table}

\paragraph{Enhanced 3D Generation with Simple Cross-Modal Fusion.}

Images and text provide complementary constraints: text offers high-level, multi-view semantic priors, while images provide precise cues on style, texture, geometry, and color. To verify this complementarity, we conduct a simple fusion experiment. At inference time, we take two pre-trained rectified-flow models, one image-conditioned and the other text-conditioned, and directly average their predicted velocity fields at each denoising step to form a joint text-image baseline, which we call SimFusion.

As shown in~\cref{tab:view_0_res}, this naive fusion already outperforms both image-only and text-only models by a large margin (82.40 FD vs. 125.93 and 145.06), suggesting that it preserves the semantic correctness from text while retaining the fine-grained visual cues from the image. This complementary effect motivates us to define a new task, termed \textbf{Text--Image Conditioned 3D Generation}.

\paragraph{Problem Formulation.}
In text-image conditioned 3D generation, the model should jointly adopt a visual exemplar (image) and a semantic description (text) to generate a coherent 3D object. Formally, given an image condition $\mathbf{I}$ and a text condition $\mathbf{T}$, the goal is to model the conditional distribution $p(\mathcal{O} \mid \mathbf{I}, \mathbf{T})$, where $\mathcal{O}$ denotes the target 3D representation (\emph{e.g.}, mesh, 3DGS, or radiance field). This task requires the model to (i) satisfy the semantics in $\mathbf{T}$ and (ii) match the view-specific appearance constraints in $\mathbf{I}$.

\section{Method}
\label{sec:method}
In this section, we introduce \textbf{TIGON} (Text–Image conditioned GeneratiON) as a baseline for our proposed task.

\begin{figure}
    \centering
    \includegraphics[width=0.9\linewidth]{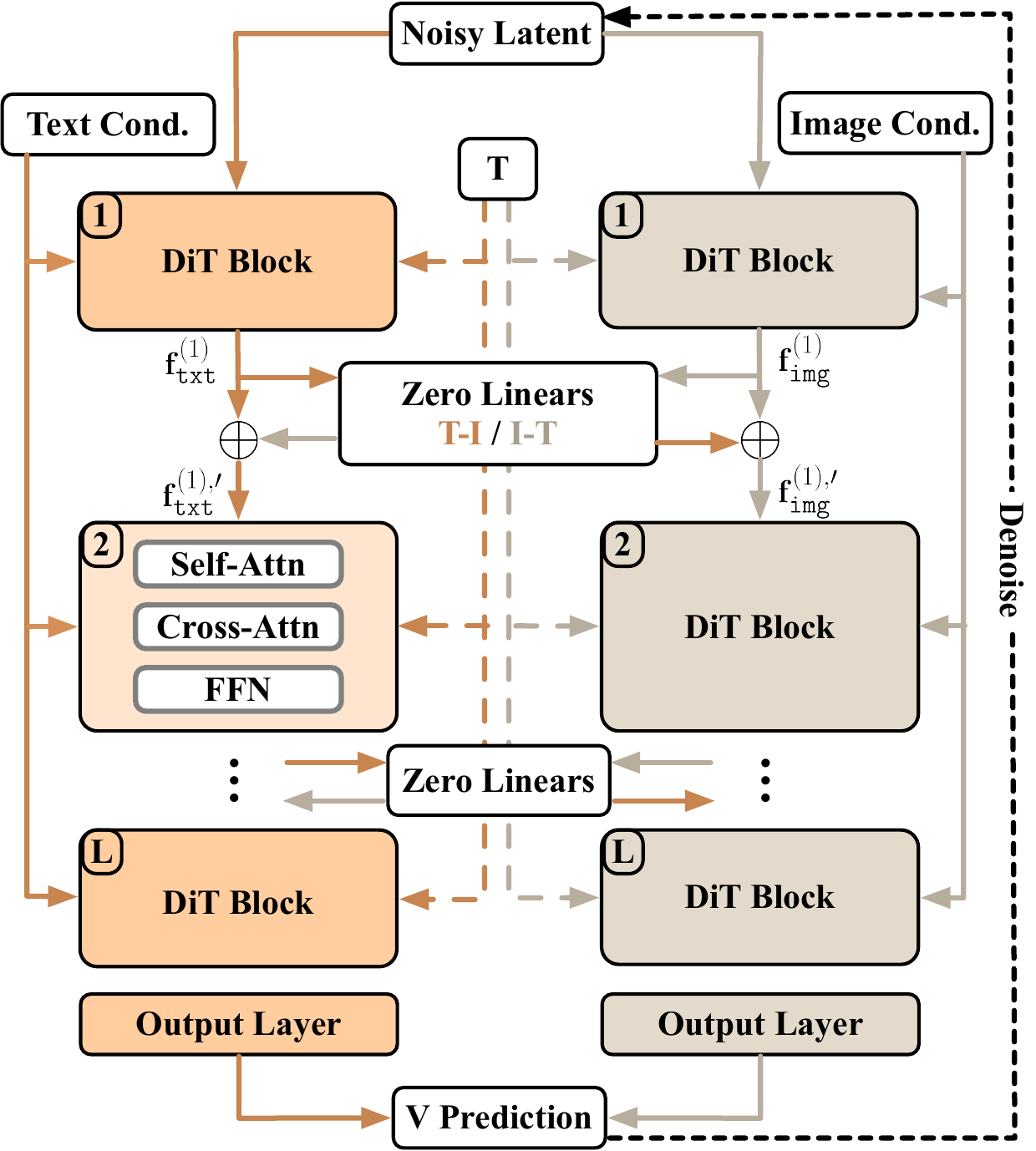}
    \caption{TIGON employs a dual-branch architecture, with a text-conditioned DiT (left) and an image-conditioned DiT (right). Paired blocks exchange features via cross-modal bridges (``Zero Linears''). At each denoising step, two predictions are averaged to produce the velocity field $\mathbf{v}$. $T$ denotes the denoising timestep.}
    \label{fig:pipe}
\end{figure}

\subsection{Overall Pipeline}
\label{sec:overall}
As shown in~\cref{fig:pipe}, \textbf{TIGON} uses two parallel branches (image- and text-conditioned) that exchange features via zero-initialized cross-modal bridges between corresponding DiT blocks. At each denoising step, their predictions are averaged to produce the final velocity field $\mathbf{v}$.

\subsection{Dual-Branch Backbone}

TIGON adopts a dual-branch backbone because image and text provide fundamentally different signals for 3D generation. Image-conditioned tokens are dense, view-grounded, and locally informative, offering explicit cues about color, texture, and fine geometry, whereas text-conditioned tokens encode sparse, abstract semantics. This asymmetry creates a granularity mismatch: for example, the concept ``tiger'' may be conveyed by a single word token but requires many image tokens to depict. Without enough data, mixing such heterogeneous token semantics within a single backbone often degrades performance. Therefore, TIGON retains two modality-specific backbones and performs fusion explicitly, preserving each branch's strengths while avoiding overly aggressive entanglement.

Each branch of TIGON is a Diffusion Transformer (DiT)~\cite{dit} with $L$ blocks. Let $\mathcal{F}_{\texttt{img}}$ and $\mathcal{F}_{\texttt{txt}}$ denote the image- and text-conditioned branches. Given latent $\tilde{\mathbf{z}}$, time step $t$, and condition (image $\mathbf{I}$ or text $\mathbf{T}$), each branch predicts a velocity field:
\begin{equation}
\begin{aligned}
    \mathbf{v}_{\texttt{img}} &= \mathcal{F}_{\texttt{img}}\!\left(\tilde{\mathbf{z}},\, t,\, \mathbf{I}\right),\\
    \mathbf{v}_{\texttt{txt}} &= \mathcal{F}_{\texttt{txt}}\!\left(\tilde{\mathbf{z}},\, t,\, \mathbf{T}\right).
\end{aligned}
\end{equation}
Both branches are pretrained in the same latent space introduced in~\cref{sec:trellis}, which allows simple additive fusion of the predicted velocities.

\subsection{Early-Fusion Strategy}
Simply averaging the final predictions of two rectified flow models is often sub-optimal: without explicit interaction, the branches can diverge and destructive averaging degrades detail and consistency. We therefore assign a \textbf{cross-modal bridge} at every backbone block for early, fine-grained cross-modal feature fusion.

Let the image- and text-conditioned branches each have \(L\) blocks. Denote by \(\mathbf{f}^{(i)}_{\texttt{img}}\) and \(\mathbf{f}^{(i)}_{\texttt{txt}}\) the output of the \(i\)-th block (\(i{=}1,\dots,L\)). We insert learned linear projections $\mathcal{P}^{(i)}_{\texttt{txt}\rightarrow\texttt{img}}$ and $\mathcal{P}^{(i)}_{\texttt{img}\rightarrow\texttt{txt}}$ to inject information across branches. The inputs to the \((i{+}1)\)-th blocks are:
\begin{equation}
\label{eq:feature-switch}
\begin{aligned}
\mathbf{f}^{(i),\prime}_{\texttt{img}} &= \mathbf{f}^{(i)}_{\texttt{img}} \;+\; \mathcal{P}^{(i)}_{\texttt{txt}\rightarrow\texttt{img}}\!\big(\mathbf{f}^{(i)}_{\texttt{txt}}\big),\\
\mathbf{f}^{(i),\prime}_{\texttt{txt}} &= \mathbf{f}^{(i)}_{\texttt{txt}} \;+\; \mathcal{P}^{(i)}_{\texttt{img}\rightarrow\texttt{txt}}\!\big(\mathbf{f}^{(i)}_{\texttt{img}}\big).
\end{aligned}
\end{equation}

\paragraph{Stability via Zero-Initialization}
Inspired by ControlNet~\cite{controlnet}, to maintain training stability at the start of joint training, all cross-modal bridges are zero-initialized. Consequently, \(\mathbf{f}^{(i),\prime}_{\texttt{img}}=\mathbf{f}^{(i)}_{\texttt{img}}\) and \(\mathbf{f}^{(i),\prime}_{\texttt{txt}}=\mathbf{f}^{(i)}_{\texttt{txt}}\) initially, and gradients progressively ``open'' these gates, learning when (and how much) to exchange information at each depth.

\subsection{Late-Fusion Strategy}
\label{sec:late_fusion}
We adopt a simple prediction-averaging scheme. Given the outputs of the text and image branches, $\mathbf{v}_{\texttt{txt}}$ and $\mathbf{v}_{\texttt{img}}$, the fused prediction at each denoising step is
\begin{equation}
    \mathbf{v} = \tfrac{1}{2}\big(\mathbf{v}_{\texttt{txt}} + \mathbf{v}_{\texttt{img}}\big).
\end{equation}

\paragraph{Why Is a Sophisticated Fusion Strategy Unnecessary?}
Early fusion with cross-modal bridges and end-to-end fine-tuning enable each branch to implicitly condition on both modalities, so any potential benefit of dynamic, modality-weighted fusion can be absorbed into the branch parameters (\emph{i.e.}, by reparameterization during training). To validate this, we compare against two learnable late-fusion variants: (i) a weight-prediction module that outputs a scalar for linear mixing; and (ii) an additional cross-modal attention block. Both yield at most marginal gains while introducing extra parameters and training variance. See~\cref{sec:ablation} for quantitative results; architectural details are in the supplementary material.

\subsection{Training Strategy}
The training of TIGON is a two-stage process. The two branches are first pre-trained separately on their respective modalities to ensure balanced learning. Then, the zero-initialized cross-modal bridges are trained, and all model parameters are jointly fine-tuned.


To preserve unimodal generation capability, we apply condition dropout: during training, the image and text conditions are independently dropped with probability 0.5. This produces a uniform mixture over four regimes—25\% unconditional (for CFG~\cite{cfg}), 25\% text-only, 25\% image-only, and 25\% text+image. Consequently, TIGON learns to handle \textbf{free-form conditioning} at inference, supporting text-only, image-only, or joint text–image inputs.

%% file: 4_experiments.tex
\section{Experiments}
\label{sec:experiments}
We first describe the implementation details, datasets, and evaluation protocol, then report quantitative and qualitative results and ablations.
\subsection{Implementation Details}
TIGON is implemented in PyTorch~\cite{pytorch} on top of TRELLIS and UniLat3D. 
We use the released UniLat3D checkpoint for the image branch. 
For the text branch, we reuse the UniLat3D backbone, replace its DINO-based condition encoder with a CLIP text encoder, and train from scratch for $1{,}000{,}000$ iterations with batch size $256$ and learning rate $1\times10^{-4}$.
We then jointly fine-tune both branches and the cross-modal bridges for $50{,}000$ iterations with learning rate $1\times10^{-5}$ in BF16 on $64$ NVIDIA A800 GPUs, using DeepSpeed ZeRO-2~\cite{deepspeed} and FlashAttention~\cite{flashattention}. 

\begin{table*}[t]
\centering
\small
\caption{Quantitative results on Toys4K (left) and UniLat1K (right). ``Cond.'' denotes conditioning modality (``I'': image, ``T'': text), and ``Rep.'' denotes output representation (``M.'': mesh, ``GS'': 3DGS).}
\label{tab:toys4k}
\setlength{\tabcolsep}{5.6pt}
\begin{tabular}{lcc|cccc|cccc}
\toprule
\multirow{2}{*}{\textbf{Model}} & \multirow{2}{*}{\textbf{Cond.}} & \multirow{2}{*}{\textbf{Rep.}} &
\multicolumn{4}{c|}{\textbf{Toys4K}} &
\multicolumn{4}{c}{\textbf{UniLat1K}} \\
 & & &
\textbf{CLIP$\uparrow$} & \textbf{FD$_{\text{DINOv2}}\downarrow$} & \textbf{ULIP$\uparrow$} & \textbf{Uni3D$\uparrow$} &
\textbf{CLIP$\uparrow$} & \textbf{FD$_{\text{DINOv2}}\downarrow$} & \textbf{ULIP$\uparrow$} & \textbf{Uni3D$\uparrow$} \\
\midrule
TripoSR~\cite{triposr}                & I & M.  & 83.14 & 596.44 & 27.37 & 24.38 & 83.37 & 652.27 & 25.90 & 23.96 \\
TRELLIS                               & I & M.  & 89.09 & 171.44 & 39.97 & 35.61 & 89.40 & 233.53 & 39.37 & 35.40 \\
TRELLIS                               & I & GS  & 90.50 &  98.75 &   -   &   -   & 90.83 & 177.20 &   -   &   -   \\
Step1X-3D$^{\dagger}$~\cite{step1x}   & I & M.  & 84.77 & 361.44 & 34.15 & 30.04 & 85.36 & 402.25 & 33.62 & 30.38 \\
Hunyuan3D-2.1$^{\dagger}$~\cite{hunyuan3d2025hunyuan3d2.1} & I & M.  & 87.57 & 171.91 & 40.22 & 35.70 & 87.27 & 249.66 & 39.58 & 35.54 \\
Stable3DGen~\cite{ye2025hi3dgen}      & I & M.  &   -   &   -    & 35.52 & 31.76 &   -   &   -    & 35.26 & 32.08 \\
Direct3D-S2~\cite{direct3ds2}         & I & M.  &   -   &   -    & 33.47 & 29.29 &   -   &   -    & 32.74 & 29.44 \\
UniLat3D                              & I & M.  & 91.85 & 109.68 & 40.32 & 35.75 & 90.00 & 205.72 & 39.60 & 35.49 \\
UniLat3D                              & I & GS  & 91.20 &  85.30 &   -   &   -   & 91.40 & 155.99 &   -   &   -   \\
\rowcolor{rowgray} TIGON (Ours)       & I & GS  & 91.40 &  84.62 &   -   &   -   & 91.64 & 153.79 &   -   &   -   \\
\midrule
TRELLIS                               & T & M.  & 87.15 & 182.42 & 37.41 & 33.59 & 85.90 & 316.05 & 36.55 & 33.23 \\
TRELLIS                               & T & GS  & 86.30 & 148.21 &   -   &   -   & 84.75 & 288.55 &   -   &   -   \\
UniLat3D                              & T & M.  & 87.03 & 179.95 & 36.14 & 32.35 & 85.29 & 313.85 & 35.34 & 32.08 \\
UniLat3D                              & T & GS  & 86.14 & 154.88 &   -   &   -   & 85.75 & 282.36 &   -   &   -   \\
\rowcolor{rowgray} TIGON (Ours)       & T & GS  & 86.77 & 152.34 &   -   &   -   & 86.42 & 273.97 &   -   &   -   \\
\midrule
SimFusion (Ours)       & I+T & GS  & 91.95 &  66.78 &  - & -  & 92.09 & 136.97 & - & - \\
\rowcolor{rowgray} TIGON (Ours)       & I+T & M.  & \textbf{92.97} &  80.77 & \textbf{41.36} & \textbf{36.68} & 90.91 & 176.69 & \textbf{40.95} & \textbf{36.74} \\
\rowcolor{rowgray} TIGON (Ours)       & I+T & GS & 92.33 & \textbf{61.59} &   -   &   -   & \textbf{92.42} & \textbf{130.08} &   -   &   -   \\
\bottomrule
\end{tabular}
\begin{flushleft}
$^{\dagger}$ Using non-public training data.
\end{flushleft}
\end{table*}

\subsection{Datasets and Evaluation Protocol}
\label{sec:eval_metric}
\noindent\textbf{Training.} TIGON is trained on \textbf{TRELLIS-500K}. Please refer to the supplement for more details about this dataset.

\noindent\textbf{Evaluation.} We evaluate on two test sets:
\textbf{Toys4K} contains about 4K high-quality 3D objects from 105 categories and is widely used by TRELLIS and UniLat3D. \textbf{UniLat1K} is a harder 1K-object benchmark curated by UniLat3D, containing 500 high-quality Sketchfab assets and 500 Toys4K samples.


\noindent\textbf{Metrics and Protocol.}
We use four metrics: CLIP~\cite{clip}, FD$_{\text{DINOv2}}$, ULIP~\cite{ulip}, and Uni3D~\cite{uni3d}. CLIP and FD$_{\text{DINOv2}}$ are computed from renderings of generated and ground-truth objects, while ULIP and Uni3D measure image--point-cloud alignment and are thus only reported for mesh outputs. To test robustness to viewpoint informativeness, each case is conditioned on three reference views (front, top, and bottom) instead of ideal views. We use public checkpoints for prior methods and re-evaluate them under this unified protocol; full metric definitions and rendering settings are provided in the supplement.

\subsection{Quantitative Results}



Results on Toys4K and UniLat1K are reported in Table~\ref{tab:toys4k}. 
We evaluate TIGON under three conditioning regimes, \emph{i.e.}, text-only, image-only, and text–image. 
TIGON is competitive in both single-modality settings, while the largest gains appear under text–image conditioning, showing effective use of complementary signals.

\begin{figure*}
\centering
    \includegraphics[width=\textwidth]{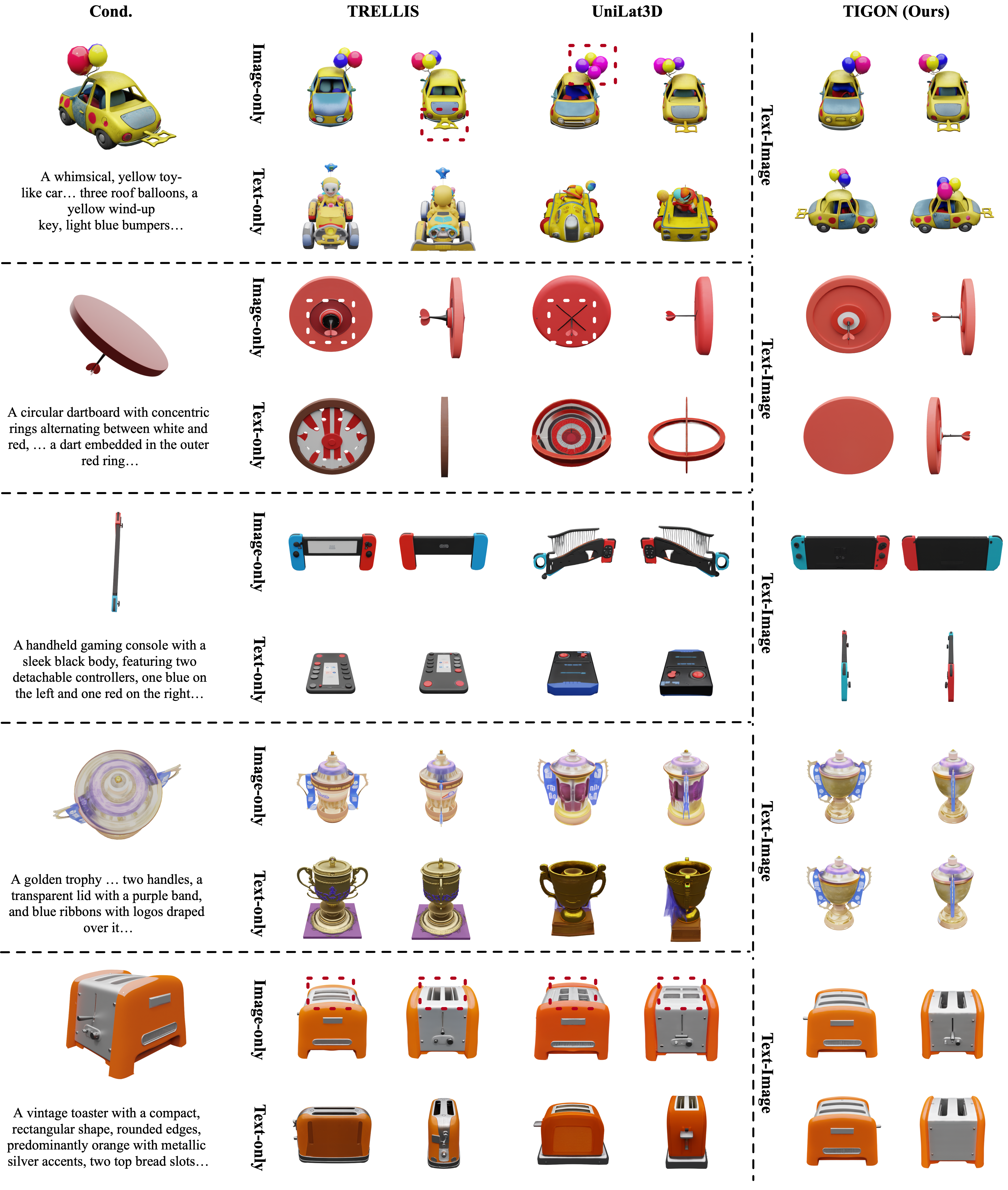}
    \caption{Qualitative comparison on Toys4K and UniLat1K against image-only and text-only variants of TRELLIS and UniLat3D. Dashed boxes mark artifacts from prior methods. Image-only models respect the reference view but must hallucinate unseen regions, while text-only models lack pixel-aligned cues and often produce low-fidelity geometry and appearance. Full prompts are provided in the supplement.}
    \label{fig:main_comp}
\end{figure*}

\subsection{Qualitative Results}
As shown in~\cref{fig:main_comp}, we compare TIGON with image-only and text-only variants of TRELLIS and UniLat3D.


\noindent\textbf{Image-Only Conditioning.}
With only one reference view, shape and appearance remain under-constrained, so image-only models must hallucinate unseen regions and often deviate from user intent. For example, given a top view of a trophy, TRELLIS/UniLat3D capture the overall style but fail to reconstruct a faithful trophy. Even with a more informative view (\emph{e.g.}, the toaster in the last row), they still produce distorted slots due to incomplete observation. Adding text supplies the missing semantics, so TIGON better matches both the description and the reference image.

\noindent\textbf{Text-Only Conditioning.}
Text provides high-level semantics but no pixel-aligned cues, leading to ambiguous geometry and lower visual fidelity. Introducing even a weak image cue markedly improves spatial alignment and appearance: in the third row of~\cref{fig:main_comp}, a top-view image of a game console combined with text produces much more faithful geometry than the text-only baseline. Overall, the visual results show that text and image address complementary failure modes, and that joint conditioning enables more controllable, higher-quality 3D generation.

\begin{figure*}
\centering
    \includegraphics[width=\textwidth]{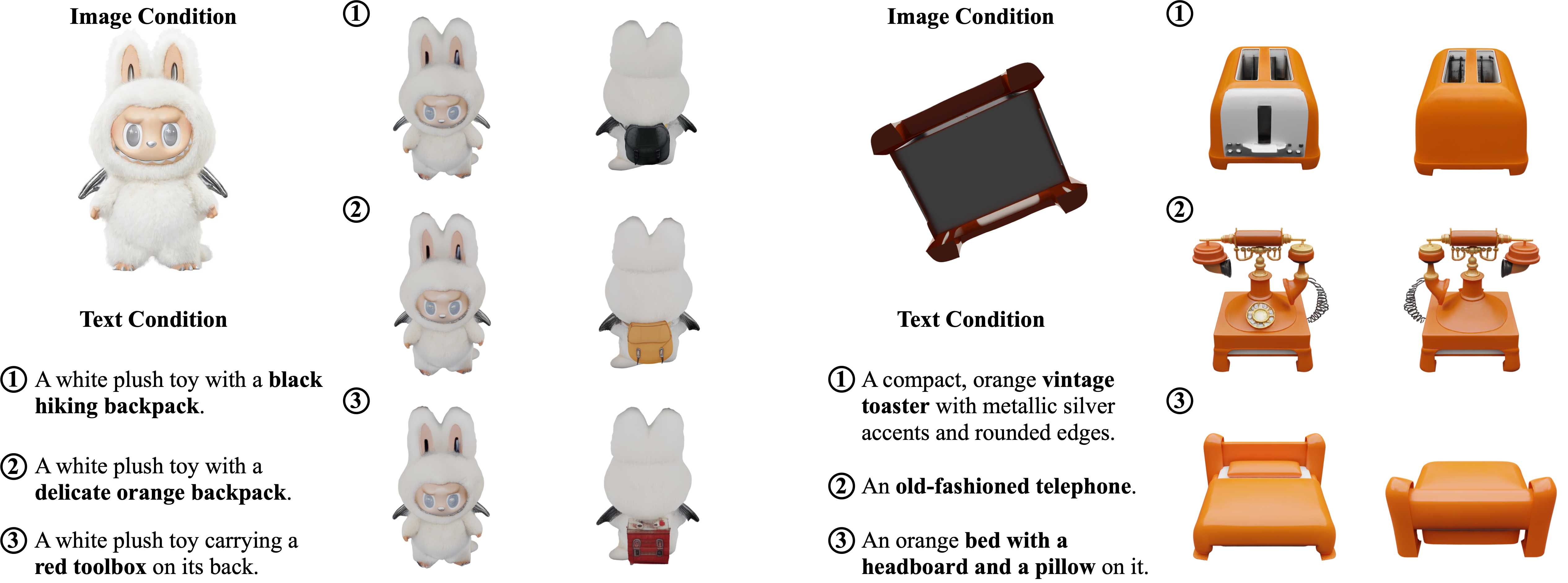}
    \caption{Controllable generation under text and image conditions.}
    \label{fig:edit}
\end{figure*}

\begin{figure}
    \centering
    \includegraphics[width=\linewidth]{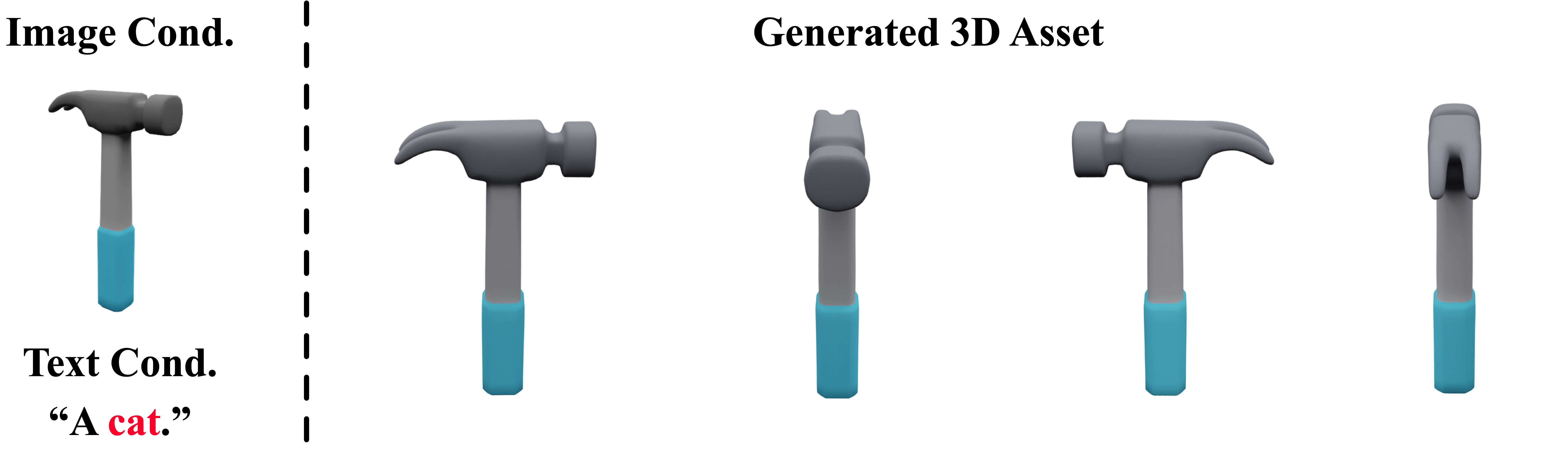}
    \caption{Generation with conflicting text-image conditions.}
    \label{fig:contradict}
\end{figure}


To further demonstrate TIGON's controllability, we fix the input image and vary the text prompt to obtain diverse 3D outputs. As shown in~\cref{fig:edit}, when the image is highly informative (\emph{e.g.}, a distinctive character), TIGON preserves identity while adjusting fine-grained attributes according to the text; when the image is ambiguous (\emph{e.g.}, a bottom view of a toaster), it relies more on text and can generate semantically different objects, such as a telephone or a bed. This combination of pixel-level alignment and semantic control offers greater flexibility than single-modality methods. We further observe in~\cref{fig:contradict} that when the image and text explicitly conflict, TIGON tends to follow the image if it already provides clear semantic guidance, likely because images are usually more specific and less ambiguous than text. More qualitative results are provided in the supplement.


\begin{table}[htbp]
    \centering
    \small
    \caption{Ablations on Toys4K. ``Bridges'' denotes zero-initialized cross-modal bridges; ``Sim'', ``AW'', and ``AT'' denote three fusion strategies; ``FT'' denotes joint fine-tuning. The TIGON setting is highlighted in light gray.}
    \begin{tabular}{ccccccc}
        \toprule
        \multirow{2}{*}{\textbf{Bridges}} & \multicolumn{3}{c}{\textbf{Fusion Strategy}} & \multirow{2}{*}{\textbf{FT}} & \multirow{2}{*}{\textbf{CLIP$\uparrow$}} & \multirow{2}{*}{\textbf{FD$_{\text{DINOv2}}\downarrow$}} \\
        & \textbf{Sim} & \textbf{AW} & \textbf{AT} & & & \\
        \midrule
        & \checkmark & & & & 91.95 & 66.78 \\
        & \checkmark & & & \checkmark & 92.05 & 66.04 \\
        \rowcolor{rowgray}
        \checkmark & \checkmark & & & \checkmark & 92.33 & 61.59 \\
        \checkmark & & \checkmark & & \checkmark & 92.31 & 60.90 \\
        \checkmark & & & \checkmark & \checkmark & 92.26 &  62.00\\
        \bottomrule
    \end{tabular}
    \label{tab:ablation}
\end{table}

\subsection{Ablation Study}
\label{sec:ablation}
We study each TIGON component on Toys4K.

\begin{figure}
    \centering
    \includegraphics[width=\linewidth]{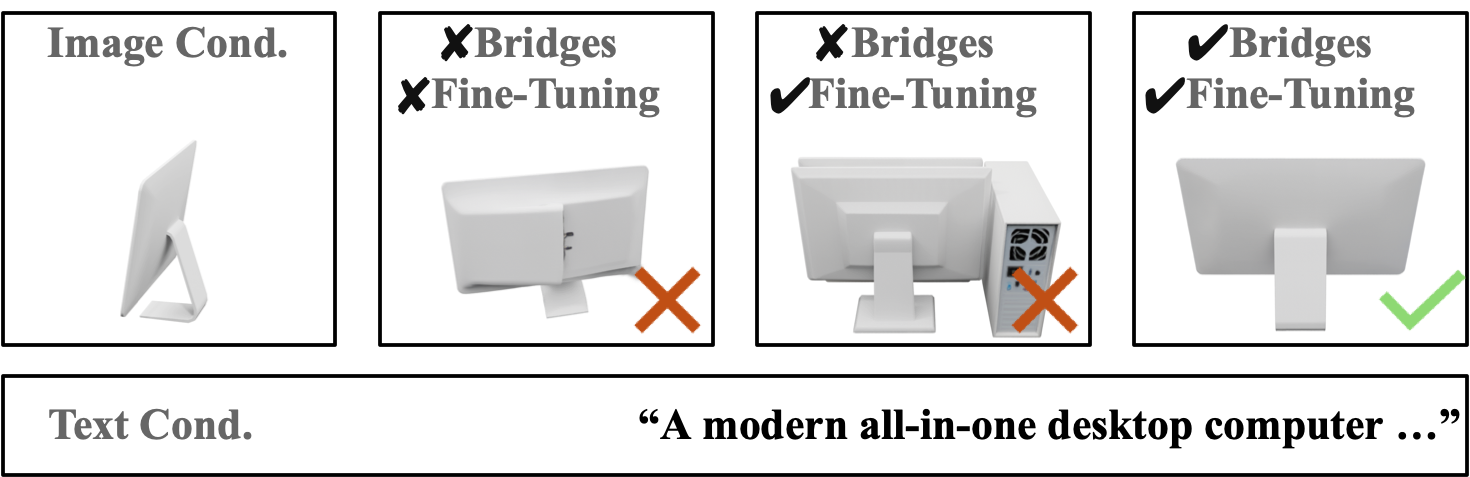}
    \caption{Effect of early fusion. Without cross-modal bridges, the two branches diverge during denoising. Full text prompt is available in the supplement.}
    \label{fig:ablation}
\end{figure}

\paragraph{Early-Fusion Strategy.}
As shown in \cref{tab:ablation}, without cross-modal bridges, joint fine-tuning of the two branches only brings marginal improvement (66.78 $\rightarrow$ 66.04 in $\mathrm{FD}_{\text{DINOv2}}$). 
Enabling cross-modal bridges yields a substantial gain (66.78 $\rightarrow$ 61.59 in $\mathrm{FD}_{\text{DINOv2}}$), underscoring the necessity of cross-modal information exchange. 
Qualitatively, \cref{fig:ablation} shows that, without cross-modal bridges, the text- and image-conditioned branches diverge during denoising, producing inconsistent or abnormal structures; with bridges, they remain aligned and produce coherent result.

\paragraph{Late-Fusion Strategy.}
As discussed in~\cref{sec:late_fusion}, a sophisticated learnable late-fusion strategy appears unnecessary. Under the same early-fusion setup and joint fine-tuning, simple averaging (\textbf{Sim}) already reaches 61.59 $\mathrm{FD}_{\text{DINOv2}}$ (Table~\ref{tab:ablation}), while adaptive weighting (\textbf{AW}) and attention-based fusion (\textbf{AT}) change this metric only slightly (60.90 and 62.00). We therefore adopt simple averaging by default.

%% file: 5_conclusion.tex
\section{Conclusion}
\label{sec:conclusion}
In this paper, we revisit conventional single-modality conditioned 3D generation and highlight a clear limitation: image-conditioned models are sensitive to viewpoint informativeness and lack control over unobserved regions, while text-conditioned models capture global intent but lack concrete visual cues. Our diagnostic study empirically shows that these two signals are complementary, which motivates us to formalize the task of \textbf{Text--Image Conditioned 3D Generation}. We further introduce \textbf{TIGON}, a simple yet effective baseline for this task. Experiments demonstrate consistent gains over strong single-modality models, and show that combining modalities yields more robust and flexible 3D generation. We hope this work will help drive future research on controllable, high-quality 3D generation.


%% file: X_suppl.tex
\clearpage
\renewcommand{\thepage}{A\arabic{page}}
\setcounter{page}{1}
\maketitlesupplementary
\appendix
\renewcommand{\thefigure}{A\arabic{figure}}
\renewcommand{\thetable}{A\arabic{table}}
\setcounter{figure}{0}
\setcounter{table}{0}

\section{Contents}
\label{sec:contents}
In this supplement, we provide the following contents:
\begin{itemize}
    \item The design of the cross-modal late-fusion module used for ablation.
    \item The dataset composition of TRELLIS-500K.
    \item The evaluation metric definitions and rendering settings.
    \item Compatibility between TIGON and TRELLIS.
    \item More qualitative results.
    \item Full text prompts used in the main paper.
\end{itemize}

\section{Design of Ablation for the Cross-Modal Late-Fusion Module}
To assess whether more sophisticated mechanisms can outperform our simple averaging strategy, we design two learnable late-fusion variants.

\begin{figure}[htbp]
    \centering
    \includegraphics[width=\linewidth]{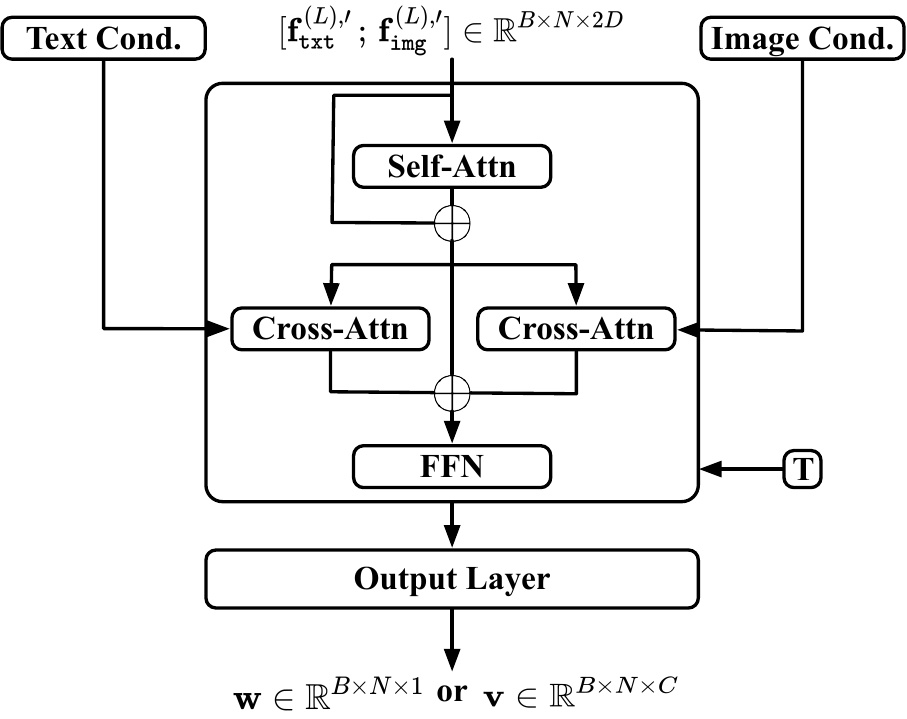}
    \caption{The adaptive fusion module used in the ablation study. 
    $D$ denotes the intermediate feature dimension, and $C$ denotes the latent output dimension.}
    \label{fig:fusion_module}
\end{figure}

A fusion module must access the current branch predictions 
$\mathbf{f}^{(L)}_{\texttt{txt}}$ and $\mathbf{f}^{(L)}_{\texttt{img}}$, 
the modality conditions $\mathbf{T}$ and $\mathbf{I}$, 
and the denoising timestep $t$. 
To integrate these inputs, we employ a dual cross-attention module. 
As shown in~\cref{fig:fusion_module}, we concatenate the two branch features along the channel dimension and obtain a fused representation:
\begin{equation}
    \mathbf{f}_{\texttt{fused}}
    = \mathcal{M}\big([\mathbf{f}_{\texttt{txt}} \,;\, \mathbf{f}_{\texttt{img}}],\, t,\, \mathbf{T},\, \mathbf{I}\big).
\end{equation}
Different output heads applied to $\mathbf{f}_{\texttt{fused}}$ produce the two fusion variants below.

\paragraph{Linear Fusion with Learnable Adaptive Weighting (AW).}
To produce element-wise fusion weights, the fusion module applies a linear projection 
$\mathbb{R}^{B\times N\times 2D} \rightarrow \mathbb{R}^{B\times N\times 1}$,
followed by a sigmoid activation to obtain 
$\mathbf{w} \in \mathbb{R}^{B\times N\times 1}$. 
The final fused prediction is computed as
\begin{equation}
    \mathbf{v}
    = \mathbf{w}\,\mathbf{v}_{\texttt{txt}}
    + (1-\mathbf{w})\,\mathbf{v}_{\texttt{img}}.
\end{equation}

\paragraph{Attention-Based Cross-Modal Fusion (AT).}
Instead of predicting fusion weights, this variant lets the fusion module directly generate modality-specific features. 
To maintain training stability, we reuse the original output projections 
$\mathcal{G}_{*} : \mathbb{R}^{D} \rightarrow \mathbb{R}^{C}$ for $* \in \{\texttt{txt}, \texttt{img}\}$. 
We split the fused representation along the channel dimension to obtain  
$\hat{\mathbf{f}}_{\texttt{txt}}, \hat{\mathbf{f}}_{\texttt{img}} \in \mathbb{R}^{B\times N\times D}$. 
Each is then projected to the latent dimension, and the final prediction is formed by summation:
\begin{equation}
    \mathbf{v}
    = \frac{1}{2}(\mathcal{G}_{\texttt{txt}}(\hat{\mathbf{f}}_{\texttt{txt}})
    + \mathcal{G}_{\texttt{img}}(\hat{\mathbf{f}}_{\texttt{img}})).
\end{equation}
Note that we omit the normalization operation for brevity.

\section{TRELLIS-500K Dataset Overview}

TRELLIS-500K is a large-scale 3D asset collection assembled in prior work~\cite{trellis}, drawing from several publicly available repositories. The dataset merges objects from Objaverse-style sources together with high-quality CAD and artist-designed assets, followed by filtering procedures to remove models with missing geometry or severely degraded textures. Each object is additionally paired with a detailed natural-language caption generated using GPT-4o~\cite{2024GPT4o}, providing consistent semantic supervision for text-driven 3D generation.

\paragraph{Objaverse-Derived ~\cite{deitke2023objaverse, deitke2024objaverse}.}
A substantial portion of TRELLIS-500K comes from higher-quality subsets of Objaverse-XL, particularly assets originating from Sketchfab (Objaverse V1) and selected GitHub contributions. These models cover a wide range of manually designed shapes, photogrammetry scans, and professionally captured artifacts. Lower-quality objects from the broader Objaverse-XL collection are excluded.

\paragraph{ABO~\cite{collins2022abo}.}
ABO contributes a set of professionally authored household product models characterized by clean topology and high-resolution materials, enriching the dataset with well-designed, manufacturable assets.

\paragraph{3D-FUTURE~\cite{fu20213d}.}
3D-FUTURE provides industrial-grade furniture models with detailed geometry and realistic textures, complementing other sources with contemporary interior designs.

\paragraph{HSSD~\cite{khanna2023hssd}.}
Assets from HSSD include indoor objects such as decorative items and furnishings originally curated for embodied AI research. These assets are structurally consistent and help broaden the dataset’s coverage of indoor categories.

Overall, TRELLIS-500K offers a curated mixture of diverse, reasonably clean 3D assets with high-quality textual descriptions, and serves as a strong large-scale dataset for training text- and image-conditioned 3D generative models.

\section{Rendering Settings and Evaluation Metric Definitions}
\subsection{Rendering Settings}
To assess generation quality, we render four reference views for each ground-truth object using cameras placed at yaw angles of ${0^\circ, 90^\circ, 180^\circ, 270^\circ}$ and a fixed pitch of $30^\circ$, all looking toward the origin with a $40^\circ$ field of view and positioned uniformly on a sphere of radius 2. We apply the same rendering protocol to the corresponding generated object to obtain its synthesized views. Image features are then extracted using the CLIP image encoder and DINOv2 to compute the CLIP similarity score and FD$_\texttt{DINOv2}$, respectively.

\subsection{Evaluation Metric Definitions}

\paragraph{CLIP.}
For each instance, we render four views for the ground-truth (GT) object and four views for the generated object, and compute the cosine similarity between every GT–generated pair, yielding a $4 \times 4$ similarity matrix. 
Since the generated objects are not canonicalized with respect to front/back orientation, we do not know the exact correspondence between GT and generated views. 
Therefore, we apply the Hungarian matching to this $4 \times 4$ matrix to find the optimal one-to-one assignment, and use the resulting matching score (average cosine similarity over the matched pairs) as the final CLIP score.

\paragraph{Fréchet Inception Distance (FID).}
To assess distributional similarity between real and generated objects, we compute a Fréchet Distance in the DINOv2 feature space. 
All GT and generated renders are embedded using a pretrained DINOv2 encoder to obtain sets of ``real'' features $\{\mathbf{x}_i\}$ and ``generated'' features $\{\mathbf{y}_i\}$. 
We estimate the empirical means and covariances $(\boldsymbol{\mu}_r, \boldsymbol{\Sigma}_r)$ and $(\boldsymbol{\mu}_g, \boldsymbol{\Sigma}_g)$ of the two distributions, and report
\begin{equation}
    \text{FD}_\texttt{DINOv2}
    = \|\boldsymbol{\mu}_r - \boldsymbol{\mu}_g\|_2^2
      + \operatorname{Tr}\!\left(\boldsymbol{\Sigma}_r + \boldsymbol{\Sigma}_g
      - 2(\boldsymbol{\Sigma}_r \boldsymbol{\Sigma}_g)^{1/2}\right),
\end{equation}
where lower values indicate that generated features more closely match the distribution of GT features.

\paragraph{ULIP and Uni3D.}
To evaluate the geometric fidelity of the generated 3D object, we employ ULIP~\cite{ulip} and Uni3D~\cite{uni3d} to measure semantic consistency between the generated 3D shape and the corresponding GT renderings. 
For each instance, we encode the four reference views using the ULIP or Uni3D image encoder, and encode the generated mesh using the corresponding point-cloud encoder applied to its vertices. 
The average cosine similarity between the image embeddings and the point-cloud embedding is reported as the ULIP or Uni3D score.

\section{Compatibility between TIGON and TRELLIS}
In this paper, we instantiate TIGON on UniLat3D, a single-stage extension of TRELLIS. However, the core idea of TIGON is not restricted to UniLat3D. To demonstrate this, we conduct an additional experiment directly on TRELLIS: we integrate the TRELLIS text and image models within the TIGON framework and evaluate the resulting model. 

Due to resource limitations, we add cross-modal bridges only to the sparse-structure flow of TRELLIS and fine-tune it, while keeping the SLAT flow as a simple fusion of the two modality branches. 

The performance is reported in Table~\ref{tab:trellis_tigon}. 
With cross-modal fusion, TRELLIS exhibits improved generation quality compared with its single-modality variants. These results highlight the potential of TIGON as a general multimodal fusion framework that is compatible with a broader class of flow-based 3D generators.  

\begin{table}[htbp]
    \centering
    \begin{tabular}{cccc}
    \toprule
         \textbf{Model} & \textbf{Cond.} & \textbf{CLIP} & \textbf{FD$_{\texttt{DINOv2}}$} \\
         \midrule
         TRELLIS & I & 90.50 & 98.75\\
         TRELLIS & T & 86.30 & 148.21\\
         TRELLIS (w/o ss-bridge) & I+T & 91.23 & 80.35\\
         TRELLIS (w/ ss-bridge) & I+T & 91.51 & 75.35\\
    \bottomrule
    \end{tabular}
    \caption{Integrating TIGON with TRELLIS. Experiment is conducted on Toys4K. We use 3D-GS as the representation. `ss-bridge' denotes the cross-modal bridge for the sparse-structure flow model.}
    \label{tab:trellis_tigon}
\end{table}

\begin{figure*}[htbp]
    \centering
    \includegraphics[width=\linewidth]{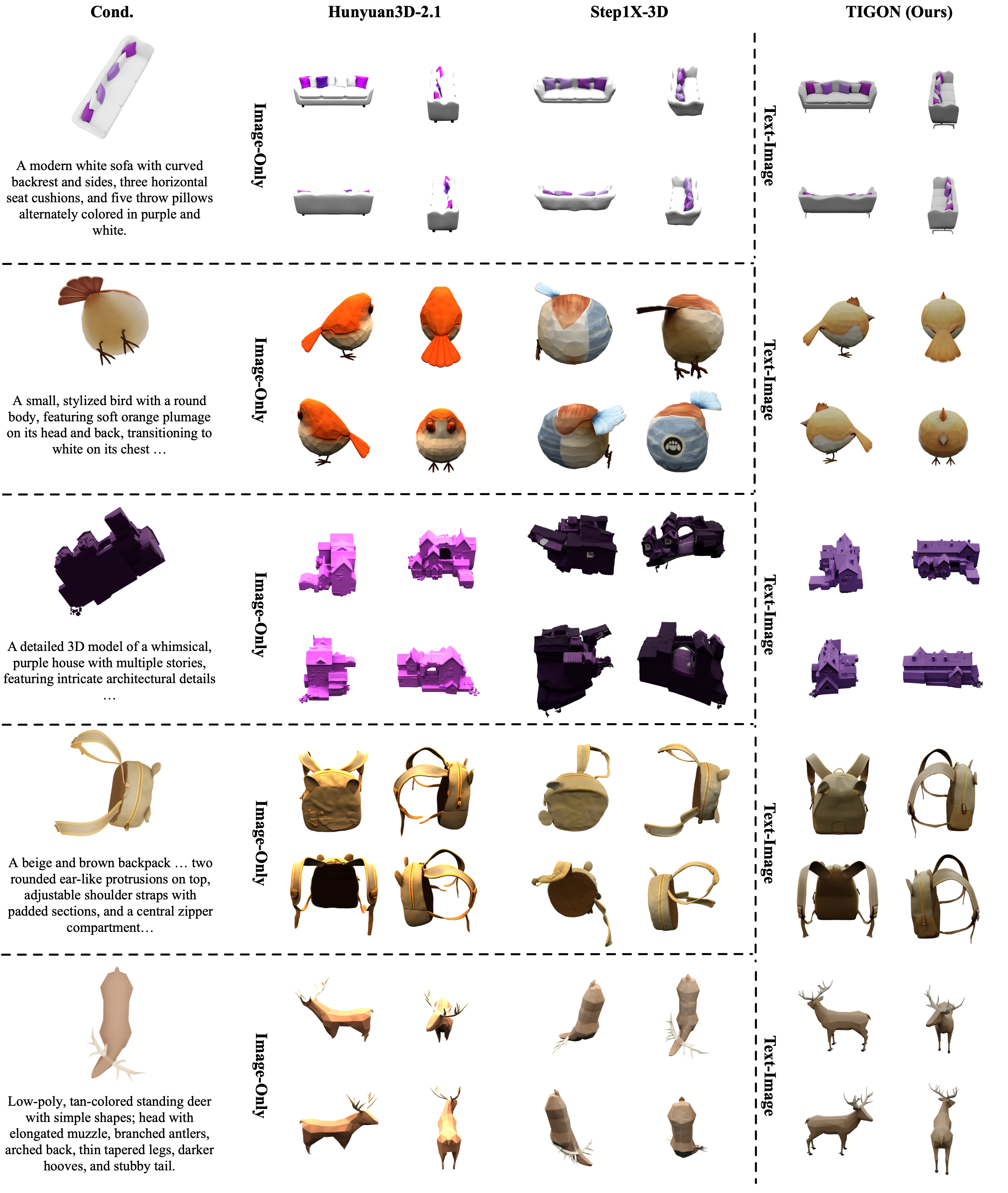}
    \caption{Meshes generated by Hunyuan3D-2.1, Step1X-3D, and TIGON. For the full text prompts, please refer to~\cref{sec:full_cap}.}
    \label{fig:mesh}
\end{figure*}

\begin{figure*}[htbp]
    \centering
    \includegraphics[width=\linewidth]{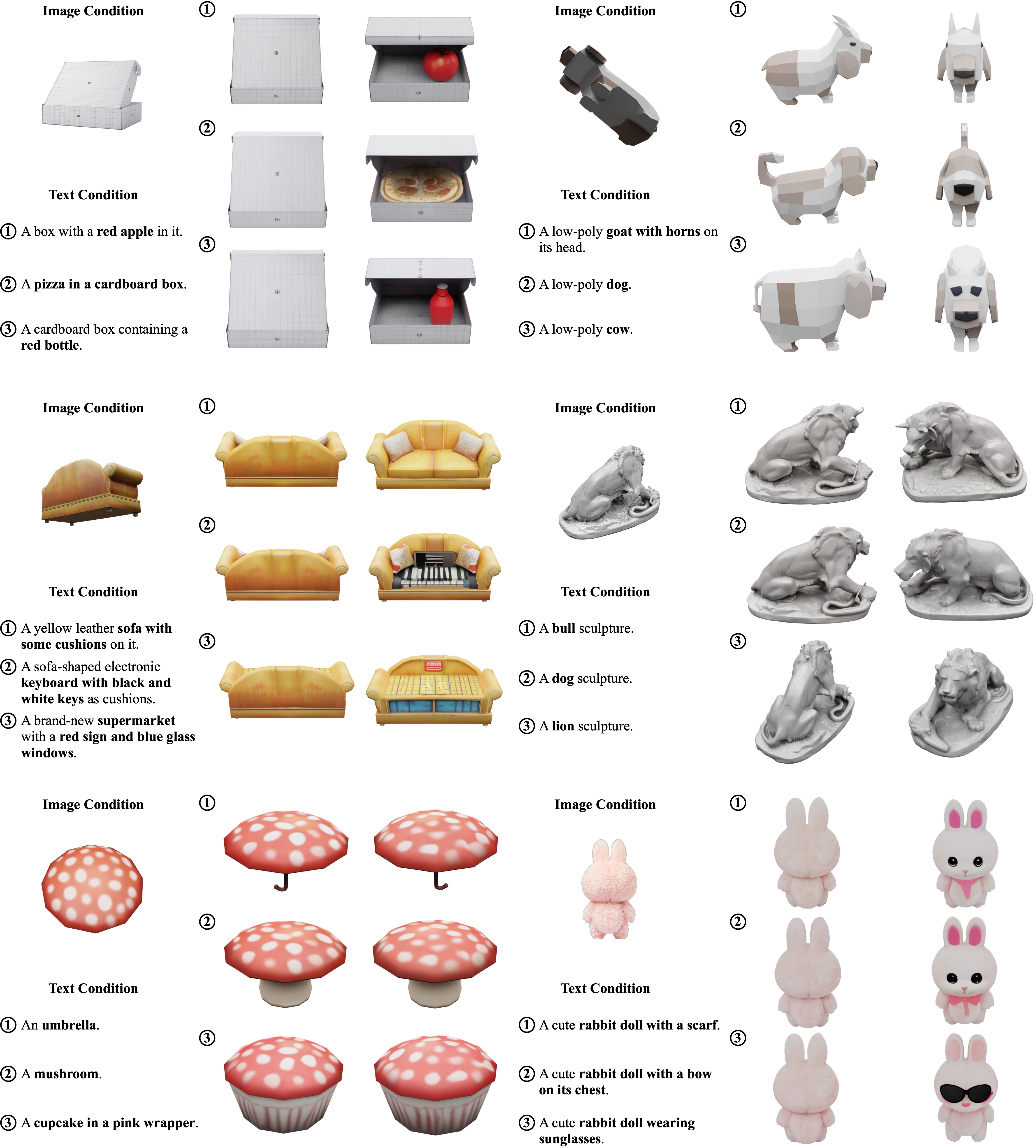}
    \caption{Additional visualization results of controllable generation with TIGON under joint text and image conditioning.}
    \label{fig:supp_edit}
\end{figure*}

\section{Additional Qualitative Results}
\paragraph{Mesh Generation.}
In the main paper, we provide visualization with 3D-GS representation. We further provide comparison results with other mesh generation methods~\cite{hunyuan3d2025hunyuan3d2.1, step1x} to demonstrate the ability of mesh generation of TIGON. 
Existing approaches exhibit a strong dependence on favorable viewpoints. 
For example, in the second row of~\cref{fig:mesh}, although the input image clearly depicts a bird, Step1X-3D fails to produce a plausible geometry. 
Similarly, in the fifth row, Hunyuan3D-2.1 cannot generate the deer’s legs. 
With explicit semantic guidance from text, TIGON successfully reconstructs these cases, producing meshes that better align with both appearance cues and object semantics.

\paragraph{Controllable Generation.}
Additional results in~\cref{fig:supp_edit} further illustrate TIGON’s controllable generation ability. By combining different condition images with different text prompts, TIGON produces diverse 3D objects while maintaining strong visual alignment, highlighting the flexibility and expressiveness enabled by joint text–image conditioning. 
We also provide videos that more fully show these controllable generation results; please refer to the supplementary materials.


\section{Full Text Prompts List}
\label{sec:full_cap}
We provide here the full text prompts that are abbreviated in our figures.

\noindent Fig. 4: From the 1st to the 5th row:
\begin{itemize}
    \item A whimsical, yellow toy-like car with a clown driver, blue side panels, and roof, red circles near the wheels, three roof balloons, a yellow wind-up key, light blue bumpers, and a weathered, vintage aesthetic.
    \item A circular dartboard with concentric rings alternating between white and red, featuring a central bullseye and a dart embedded in the outer red ring. The dart has a black shaft, a brown tip, and red fletching. The dartboard's surface appears to have a matte finish, and the lighting highlights its three-dimensional form.
    \item A handheld gaming console with a sleek black body, featuring two detachable controllers, one blue on the left and one red on the right. The blue controller has a directional pad and four action buttons labeled A, B, X, and Y. The red controller includes a joystick and additional buttons. The device has a glossy finish with visible screws and ports along its edges, indicating a portable design for gaming on the go.
    \item A golden trophy with intricate engravings and detailed textures, featuring two handles, a transparent lid with a purple band, and blue ribbons with logos draped over it. The base has multiple tiers with inscriptions, and the overall design includes reflective surfaces and polished finishes.
    \item A vintage toaster with a compact, rectangular shape, rounded edges, predominantly orange with metallic silver accents, two top bread slots, front and back panels featuring curved orange with central silver panels, inwardly curved side panels with a metallic lever and two knobs, and a wider base.
\end{itemize}

\noindent Fig. 6: A modern all-in-one desktop computer with a sleek, white stand, slightly curved white back, flat rectangular screen with rounded corners, thin black side and top bezels, and a thicker white bottom bezel.

\noindent\cref{fig:mesh}: From the 1st to the 5th row:
\begin{itemize}
\item A modern white sofa with curved backrest and sides, three horizontal seat cushions, and five throw pillows alternately colored in purple and white.
\item A small, stylized bird with a round body, featuring soft orange plumage on its head and back, transitioning to white on its chest. The bird has a short, pointed black beak, dark eyes, and delicate brown legs with clawed feet. Its tail is short and slightly fanned, matching the orange coloration of its back. The texture appears smooth and slightly fluffy, giving it a plush, cartoon-like appearance.
\item A detailed 3D model of a whimsical, purple house with multiple stories, featuring intricate architectural details such as gabled roofs, chimneys, windows with varying sizes and shapes, and a staircase leading to the entrance. The house has a textured surface resembling stone or brick, with small decorative elements like railings and a balcony. The surrounding area includes scattered debris and a small figure near the base, adding context to the scene.
\item A beige and brown backpack with a textured fabric surface, featuring two rounded ear-like protrusions on top, adjustable shoulder straps with padded sections, and a central zipper compartment. The backpack has a structured design with visible stitching details and reinforced areas around the straps and zippers.
\item Low-poly, tan-colored standing deer with simple shapes; head with elongated muzzle, branched antlers, arched back, thin tapered legs, darker hooves, and stubby tail.
\end{itemize}
